\documentclass{article}

\usepackage[preprint]{neurips_2026}
\usepackage{arydshln}
\usepackage[utf8]{inputenc}
\usepackage[T1]{fontenc}
\usepackage{hyperref}
\usepackage{url}
\usepackage{booktabs}
\usepackage{amsfonts}
\usepackage{amsmath}
\usepackage{nicefrac}
\usepackage{microtype}
\usepackage{xcolor}
\usepackage{graphicx}
\usepackage{tabularx}
\usepackage{booktabs}
\usepackage{array}
\usepackage{multirow}
\title{Reflective Agentic Physics Control for Physically Plausible Video Generation}
\newcommand{\lqr}[1]{\textcolor{black}{#1}}
\newcommand{\hjk}[1]{\textcolor{black}{#1}}
\newcommand{\yr}[1]{\textcolor{black}{#1}}
\newcommand{\maybeincludegraphics}[2][]{%
  \IfFileExists{#2}{%
    \includegraphics[#1]{#2}%
  }{%
    \fbox{%
      \begin{minipage}[c][0.25\textheight][c]{0.92\linewidth}
      \centering
      Missing figure: \texttt{\detokenize{#2}}
      \end{minipage}%
    }%
  }%
}

\author{
Qirui Li$^{1}$ \qquad
Jinkun Hao$^{1}$ \qquad
Yibo Li$^{1}$ \qquad
Ran Yi$^{1}$ \qquad
Paul L. Rosin$^{2}$ \qquad
Yu-Kun Lai$^{2}$\\
$^{1}$Shanghai Jiao Tong University
\qquad
$^{2}$Cardiff University
}

\begin{document}

\maketitle
\vspace{-0.3in}
\begin{figure*}[htbp]
    \centering
    \includegraphics[width=\textwidth]{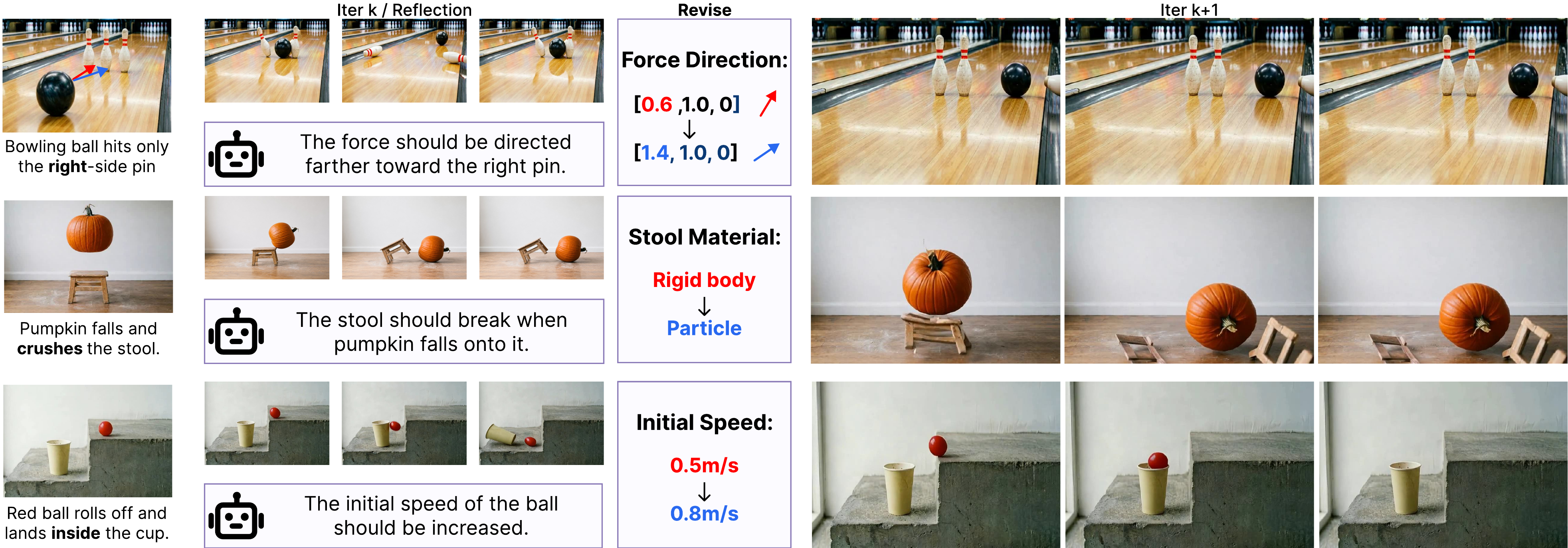}
    \caption{ We introduce PhysAgent, a reflective agentic physics control framework. Given a single input image and a physical event prompt, our method constructs a simulatable scene and iteratively diagnoses reconstruction and simulation failures. The resulting videos preserve the visual scene while satisfying fine-grained physical requirements, including force direction, material response, and object-to-object interactions.
    }
    \label{fig:teaser}
\end{figure*}

\begin{abstract}

Recent advances in physics-grounded video generation leverage physics simulation as a physical prior to guide video synthesis toward physically plausible outcomes. The simulation process is controlled by physical specifications, which are typically generated by a vision-language model in a single pass. Such one-shot prediction often fails to accurately translate user intent into executable simulations, particularly for fine-grained object dynamics, complex motion trajectories, and temporally structured interactions.
In this paper, we propose PhysAgent, a reflective agentic framework that closes the loop among physical program generation, physics simulation, stage-specific verification, and targeted program repair. Beyond improving the control of coupled physical parameters, our framework enables the agent to progressively realize complex trajectories, multi-stage interactions, and precise event outcomes by treating each physical program as an executable hypothesis. In addition, we design a set of physics-control APIs to support more stable and complex motion behaviors. Extensive experiments demonstrate that PhysAgent produces more physically plausible videos, achieves better prompt alignment, and generalizes more effectively across diverse physical scenarios. The project page is \url{https://iapple233.github.io/PhysAgent}.
\end{abstract}

\section{Introduction}

\hjk{Recent video generation methods have achieved remarkable progress in visual fidelity, temporal coherence, and controllability by extending diffusion models from images to videos}
\citep{ho2022video,ho2022imagen,singer2022makeavideo,
blattmann2023videoldm,blattmann2023stable,guo2024animatediff,bartal2024lumiere}.
\hjk{However, despite their impressive visual quality, these models often fail to produce physically plausible dynamics.}
Objects may slide without contact, respond to forces in the wrong
direction, or ignore material-dependent behavior.
These failures are especially salient in image-to-video generation, where a
single input image must be transformed into a plausible future under a
user-specified physical event.

\hjk{The central challenge extends beyond physical plausibility to fine-grained control over physical dynamics. Given a user-specified event, the system should identify the relevant objects, reconstruct a simulatable scene, infer material and contact properties, and generate temporally coherent motions that satisfy the intended physical outcome.}
Recent physics-grounded
video methods address this challenge by introducing physical intermediate
representations into generation. PhysCtrl \citep{wang2025physctrl} learns physics-conditioned point
dynamics for controllable generation; PhysGen3D \citep{chen2025physgen3d}
reconstructs an interactive 3D world from a single image and simulates
physical dynamics; RealWonder \citep{liu2026realwonder} translates
simulated video into visual priors for real-time
action-conditioned video generation. 

However, these physics-grounded methods exhibit limited robustness. This brittleness stems from the need to infer a structured simulator specification, which is typically predicted in a single pass by a VLM. The specification jointly determines scene reconstruction and subsequent physical simulation, and therefore contains many tightly coupled factors. This is inherently challenging because the simulator specification consists of many tightly coupled factors. Although each individual factor may appear reasonable in isolation, their combined effect can still produce a rollout that fails to satisfy the intended event. Therefore, the correctness of a simulator specification cannot be determined from the specification itself, but must be evaluated through execution. The system needs to execute a candidate specification, compare the resulting motion with the user instruction, and identify which part of the specification should be revised. This makes single-pass VLM prediction inherently brittle and motivates an execution-aware formulation of physics-grounded video generation.

\hjk{Building on this observation, we introduce \emph{PhysAgent}, a closed-loop agentic framework for physical simulation control. PhysAgent treats each physical specification as an executable control program. 
Given an input image and a user-specified event, PhysAgent organizes this loop around a program generator, a stage-specific verifier, and a planner. The generator constructs the physical control program, the simulator produces visual evidence, the verifier identifies reconstruction or dynamics failures, and the planner converts each diagnosis into a targeted program edit.}
In this way, physical control is transformed from single-pass parameter prediction into feedback-driven refinement.

Concretely, our framework represents each candidate as a physical control program factorized into scene state and event program. The scene state specifies grounded objects, their physical attributes, support geometry, and camera parameters recovered from the
input image. The event program is instantiated through a compact action API that supports linear and angular velocity initialization, temporally scheduled force and torque actuation, disturbance injection, pose assignment, and object fixation. The interface reduces the need for the agent to directly synthesize brittle, simulator-specific numerical commands, while retaining explicit control over actuation type, timing, and duration. The program is then executed through a progressive
validation schedule: \lqr{scene reconstruction is checked before diagnostic rollout, and the diagnostic
rollout is checked before video synthesis.} Accepted decisions are stored in a
compact memory so that later revisions preserve what has already been verified.
A verified rollout is finally rendered as a structured motion prior for video
synthesis.

Our contributions are summarized as follows:
\begin{itemize}
    \item \hjk{We formulate physically plausible video generation as an execution-aware physical simulation control problem, where simulator specifications are treated as executable hypotheses rather than one-shot predictions.}
    \item We propose a reflection-based agentic framework that closes the loop between VLM-based configuration generation, physics simulation, reflective verification, and parameter refinement.
    \item We design a semantic physics-control interface that enables more flexible, compositional, and expressive motion specification beyond simple initial-force or collision-based controls.
    \item Experiments show that the proposed framework improves physical plausibility, prompt alignment, and generalization for complex physics-grounded video generation scenarios.
\end{itemize}

\section{Related Work}

\noindent\textbf{Physics-Grounded Video Generation.}
Modern video diffusion models have achieved strong visual fidelity
\citep{ho2022video, ho2022imagen, blattmann2023stable}, but their dynamics are
still largely learned from data rather than governed by explicit physical constraints.
Recent physics-aware methods address this by injecting physical priors into video
generation. 
PhysGen uses rigid-body simulation to guide image-to-video generation under force and torque controls \citep{liu2024physgen}; 
PhysCtrl learns generative point-trajectory dynamics conditioned on physical parameters
\citep{wang2025physctrl}; 
PhysGen3D reconstructs a camera-centric interactive 3D scene from a single image for physics-based simulation \citep{chen2025physgen3d}; 
PhysVid introduces local physics-aware conditioning for generative video models \citep{pathak2026physvid}; 
and RealWonder connects single-image 3D reconstruction, physics simulation, optical-flow/RGB priors, and a distilled video generator for real-time action-conditioned video \citep{liu2026realwonder}.
These works show that physical intermediates improve
motion controllability, but they largely assume that the physical configuration
is already correct or can be specified in one pass. Our focus is orthogonal: we
study how a VLM agent can \emph{arrive at} a usable physical configuration by
iteratively executing the simulator, inspecting intermediate artifacts, and
repairing the symbolic controls before video synthesis.

\noindent\textbf{Execution-Grounded Multimodal Agents.}
Language agents have evolved from text-only reasoning to systems that act in
external environments, including web interfaces, software repositories, and
embodied simulators \citep{yao2022react, shinn2023reflexion,
madaan2023selfrefine, wang2023voyager}. Visual environments are more challenging
than text environments because failures are often visible only after executing an
action and inspecting its rendered consequence. Visual programming methods such
as ViperGPT and Visual Programming compose code over perception modules
\citep{suris2023vipergpt, gupta2023visual}, while recent graphics agents such as
BlenderGym, SceneCraft, and VIGA use executable graphics environments to support
iterative visual synthesis and editing \citep{gu2025blendergym,
hu2024scenecraft, yin2026viga}. These works motivate execution-based
verification, but their main target is visual or geometric correctness. Our
setting adds an additional layer: the agent must verify whether a simulated
temporal process satisfies force, contact, support, and material requirements.

\noindent\textbf{Semantic Interfaces for Programmatic Control.}
Programmatic visual systems expose expressive but brittle low-level APIs. In
graphics, LLM agents often rely on structured scene programs, reusable skills, or
tool interfaces to reduce syntax errors and make edits more semantically
meaningful \citep{hu2024scenecraft, gu2025blendergym, yin2026viga}. Physics
simulation has a similar but sharper interface problem: small changes in force
duration, collision geometry, gravity direction, friction, or object anchoring
can qualitatively change the resulting motion. We therefore treat high-level
physics controls as a central part of the method rather than merely as an engineering convenience. By exposing semantic primitives for staged forces, state-dependent
impulses, fixation, release, and feedback controllers, our framework lets the
agent reason at the level of physical events while still compiling to concrete
simulator operations.

\section{Method}
\label{sec:method}


\hjk{Given a single image and an event prompt, our goal is to synthesize a video that preserves the visual appearance of the input while following physically plausible dynamics. Our method uses physics simulation as an explicit intermediate representation: it constructs an executable physical control program, runs the program in a physics engine, and uses the resulting simulated video to guide video synthesis. Sec.~\ref{Sec:pipeline} first introduces the single-pass simulation-grounded generation process and the physical program representation. Sec.~\ref{Sec:loop} then describes how we extend this process with a reflective simulation loop that validates, diagnoses, and repairs physical programs before final synthesis. Sec.~\ref{Sec:api} details the action APIs that define the editable control space for physical event generation.}

\begin{figure*}[t]
    \centering
    \includegraphics[width=\textwidth]{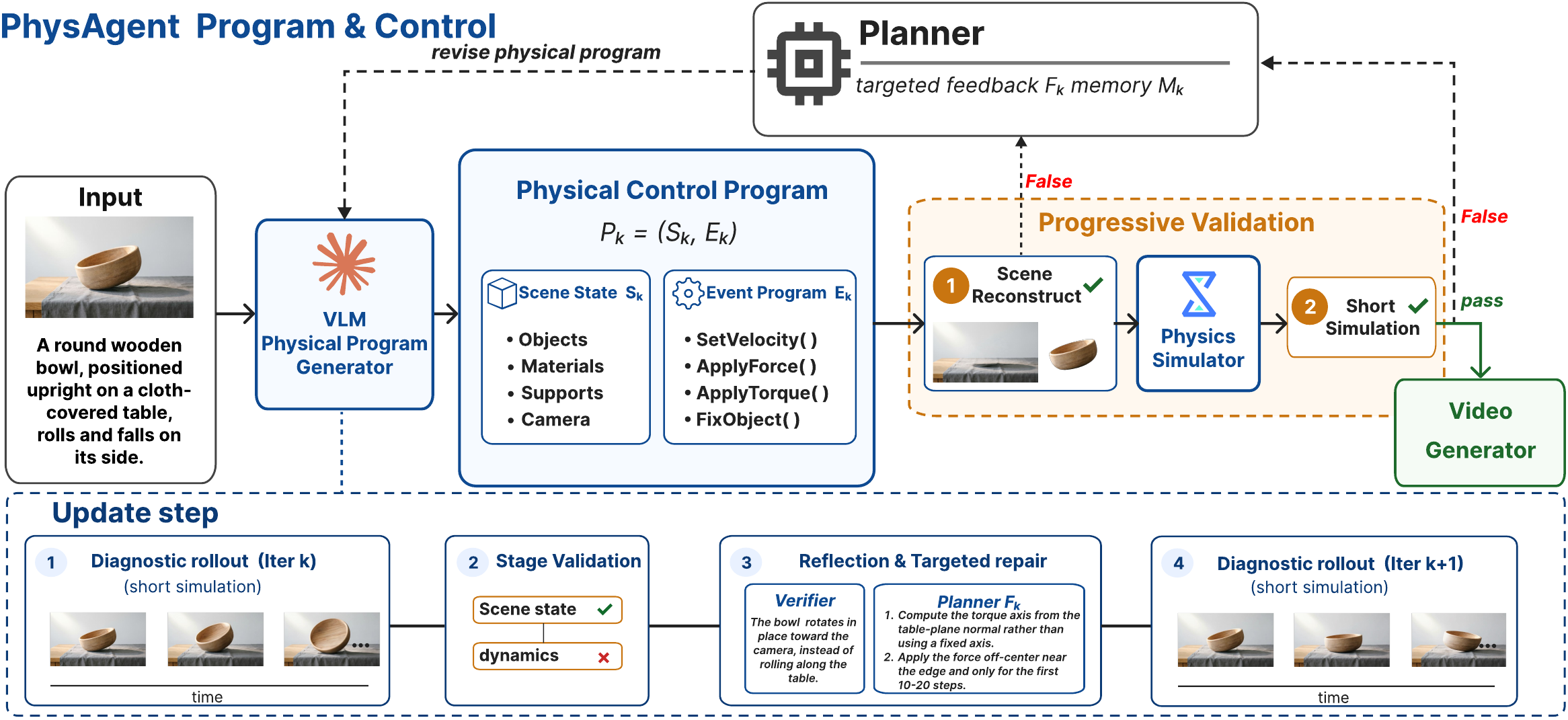}
    \caption{
        Overview of our reflective physics-control pipeline. Starting from an
        input image and event prompt, the agent infers a physical control
        program, executes it in simulation, verifies intermediate artifacts, and
        refines failed factors before using the accepted rollout as a structured
        motion prior for video synthesis.
    }
    \label{fig:pipeline}
\end{figure*}

\subsection{Simulation-Grounded Generation Pipeline}
\label{Sec:pipeline}
We first introduce a single-pass simulation-grounded generation process, which defines the basic execution path of existing physics-grounded video generation methods. Given an input image \(I\) and an event prompt \(p\), the physics simulation process generates physical dynamics and renders them into a simulated video, which serves as the conditioning signal for the video generator.

\lqr{The physics simulator generally consists of two stages: scene reconstruction and physical simulation. Each stage is controlled by a corresponding structured component: the Scene State \(S\) and the Event Program \(E\). We collectively refer to \(S\) and \(E\) as physical control programs.}

In the scene reconstruction stage, scene state \(S\) provides an editable scene description, it specifies coordinates of each object in the image, material types, densities and camera pose. The coordinates of objects will guide SAM3D~\cite{chen2026sam} in segmenting the objects and reconstructing their 3D geometry.




In the physical simulation stage, event program \(E\) specifies the target objects, the magnitudes and directions of applied forces and initial velocities, and the timing of each action. The physics simulator then executes these actions in the reconstructed scene according to E. This execution produces physical trajectories, which are rendered as a simulated video. The simulated video provides motion priors, such as optical flow, for the final video generator.

Therefore, the simulation process can be summarized as
\begin{equation}
X = \mathrm{Sim}(S, E),
\end{equation}
where the simulator \(Sim\) executes the event program \(E\) under the scene state \(S\), producing \(X\), which denotes the resulting simulated video. Since both \(S\) and \(E\) are structured programs, reconstruction errors and dynamics errors can be programmatically revised by the agent. This property provides the basis for the method introduced in the following section.

\subsection{Reflective Loop}
\label{Sec:loop}


\begin{figure*}[t]
    \centering
    \includegraphics[width=\textwidth]{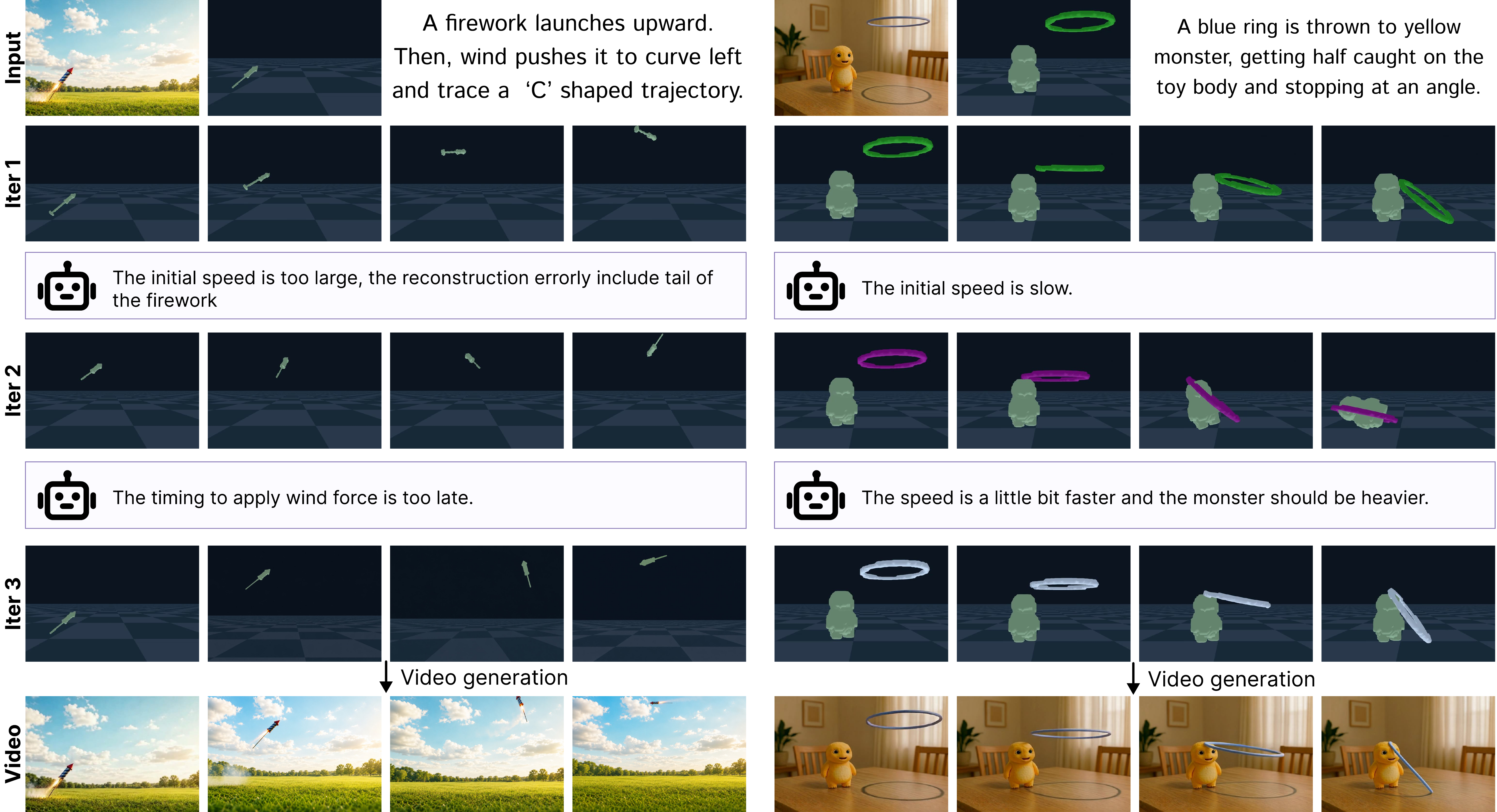}
    \caption{
        Stage-specific reflective refinement. By separating reconstruction
        feedback from simulation feedback, the verifier receives stage-specific evidence, avoiding unnecessary downstream rollouts and enabling more targeted repairs.
    }
    \label{fig:reflection}
\end{figure*}

To mitigate errors in the simulated video caused by unreasonable physical control programs in the single-pass pipeline, we introduce a feedback-driven agent system, PhysAgent. The system consists of a generator $\mathcal{G}$, a verifier $\mathcal{V}$, and a planner $\mathcal{P}$, which jointly refine the physical control programs through iterative feedback.

\noindent\textbf{\yr{Generator.}}
The physical program generator $\mathcal{G}$ first constructs an initial
physical control program from the input image $I$ and the event prompt $p$:
\begin{equation}
P_1=(S_1,E_1)=\mathcal{G}(I,p).
\end{equation}
For each subsequent round $k>1$, the generator revises the previous program
$P_{k-1}$ according to the targeted feedback $F_{k-1}$ from the planner:
\begin{equation}
P_k=(S_k,E_k)
=\mathcal{G}\!\left(P_{k-1},F_{k-1}\right).
\end{equation}
Here, $S_k$ and $E_k$ denote the revised scene state and event program
at round $k$. The feedback $F_{k-1}$ is described below.


\noindent\textbf{\yr{Verifier.}}
At round $k$, the verifier $\mathcal{V}$ examines the stage-specific
visual evidence $Z_k$ and verifies whether it satisfies the event prompt
$p$:
\begin{equation}
    A_k=\mathcal{V}(Z_k,p),
\end{equation}
where $A_k$ contains a pass/fail decision and, upon failure, a concise
diagnosis of the observed mismatch. The verifier only identifies what is
incorrect, leaving concrete parameter updates to the planner. To localize
failures and avoid unnecessary downstream execution, we adopt a
stage-specific progressive validation procedure consisting of
reconstruction validation and physical simulation validation.

1) Validation of scene reconstruction. After scene reconstruction, the verifier receives a rendered view of
the reconstructed 3D scene, together with coordinate points of objects and camera information. If reconstruction validation fails, the simulator is not invoked. Instead, the verifier produces a diagnostic assessment $A_k$ that
identifies the failed reconstruction factors and guides the planner.

2) Validation of physical simulation.
After successfully validating scene reconstruction, the event program is executed by physics simulator. It produces physical dynamics, which are rendered to an 81-frame simulated video $X_k$. To reduce the amount of visual context provided to the verifier, we sample every third frame to obtain a compact representation of the simulation.
\begin{equation}
    \tilde{X}_k = \{x_k^1, x_k^4, x_k^7, \ldots, x_k^{79}\}.
\end{equation}
The verifier then checks whether $\tilde{X}_k$ satisfies the event prompt in
terms of object trajectory, contact timing, force direction and material response. If the simulation fails, the planner generates
a targeted revision to the event program. If the diagnostic rollout passes, we compute optical flow from the
81-frame simulated video and pass it to the video generator as a motion prior for
final video synthesis.

\noindent\textbf{\yr{Planner.}}
If verification fails, the verifier $\mathcal{V}$ provides its assessment $A_k$ to the planner. Given this assessment and current memory, the planner produces a targeted revision proposal:
\begin{equation}
F_k = \mathcal{P}(A_k, M_k).
\end{equation}
where $M_k$ is the memory of the physical program, which stores the modification records of each round of the physical program. The output feedback $F_k$ is a structured edit proposal, such as reselecting an object, correcting
force direction or magnitude and modifying action timing. The generator then revises the complete physical program
conditioned on $F_k$, such as reselecting the target object, correcting the force direction or magnitude, or modifying the action timing. This divide-and-repair strategy lets the agent repair the
failed physical factor without unnecessarily changing decisions that have already
been validated.

\begin{table}[t]
\centering
\small
\setlength{\tabcolsep}{4pt}
\renewcommand{\arraystretch}{1.08}
\caption{
Action APIs used by the semantic physical-control program. These primitives
cover state initialization, force and torque actuation, disturbance injection,
pose assignment, and object fixation, enabling the agent to compose precise
physical events without directly predicting brittle low-level simulator updates.
}
\label{tab:motion_control_actions}
\begin{tabularx}{\linewidth}{
    @{}
    p{0.20\linewidth}
    p{0.31\linewidth}
    >{\raggedright\arraybackslash}X
    @{}
}
\toprule
\textbf{Action Name} & \textbf{Arguments} & \textbf{Description} \\
\midrule
SetVelocity
& name, velocity, start step
& Assign the initial velocity of an object. \\

SetAngularVelocity
& name, angular\_velocity, start step, duration
& Set initial angular velocity. \\

ApplyForce
& name, force, point, start step, duration
& Apply a force vector to an object. \\

ApplyTorque
& name, torque, start step, duration
& Apply a torque vector. \\

ApplyAngledForce
& name, magnitude, angle, start step, duration
& Apply a force at a specified angle relative to the object's current velocity. \\

ApplyDisturbance
& name, model, amplitude, start step, duration
& Inject a perturbation. \\

SetPosition
& name, position
& Specify the position of an object. \\

SetOrientation
& name, rotation
& Specify the rotational pose of an object. \\

FixObject
& name, start step, duration
& Constrain an object to remain stationary. \\
\bottomrule
\end{tabularx}
\end{table}

\subsection{Action APIs}
\label{Sec:api}

We observe that requiring VLM to directly synthesize complex low-level simulator commands often introduces unnecessary syntactic overhead and errors, which can compromise its reasoning performance. To mitigate this, we abstract low-level operations into a set of simplified Action APIs, which will be used frequently. 
We represent an event program $E$ as:
\begin{equation}
    E=\{o_j(\theta_j,\tau_j)\}_{j=1}^{J},
\end{equation}
where $o_j$ denotes a physics-control operator, $\theta_j$ denotes its input
arguments, and $\tau_j$ denotes the time interval during which the operator is
applied. The arguments specify the target object and the parameters needed by
the operator, such as linear velocity, angular velocity, force, torque, position,
orientation, start step, and action duration.

These operators cover four common types of physical control: 1) Initial
state operators set the linear or angular velocity of an object before
simulation starts. 2) Force and torque operators apply external control to
move or rotate an object. 3) Disturbance operators add time-varying effects,
such as wind or small perturbations. 4) Pose and constraint operators set
the position or orientation of an object, or keep an object fixed for a given
time. All operators are executed by the simulator, but they allow the agent to
describe physical events using clear motion-level actions rather than raw
numerical updates.

\lqr{The operators used in our implementation are summarized in Table~\ref{tab:motion_control_actions}.} Together, these operators provide a compact control vocabulary for composing
object motion in 3D space while avoiding brittle low-level simulator-specific
commands.

\section{Experiments}

\subsection{Experimental Setup}

\paragraph{Evaluation benchmark.}
\lqr{
We construct a physical-event benchmark to evaluate whether a video generation
method can realize controllable, complex, and precise physical dynamics from a
single input image. The benchmark contains 27 distinct scenes, and each scene is
paired with 2 to 4 different textual event descriptions, resulting in 80 event
cases in total. This same-scene multi-event design is important: it requires a
method to preserve the visual scene while changing only the physical event
program, so that the verification isolates physical controllability rather than
scene appearance.
}


\paragraph{Compared methods.}
We compare our method with both general video generation models and
physics-grounded generation baselines. For general image-to-video generation, we
use Wan~\citep{wan2025wan} and include Goal-Force~\citep{gillman2026goal} as a control-oriented
baseline that directly specifies a goal-directed force. In addition, we compare against recent
physics-grounded video generation methods, including
PhysCtrl~\citep{wang2025physctrl}, PhysGen3D~\citep{chen2025physgen3d} and
RealWonder~\citep{liu2026realwonder}. Our method uses the reflective
simulation loop and the semantic physical-control program described in
Sec.~\ref{sec:method}.

\paragraph{Evaluation metrics.}
We evaluate generated videos along two complementary axes. First, we follow the
VBench protocol~\citep{huang2024vbench} and report automatic perceptual-quality metrics, including visual quality, aesthetic quality, and temporal consistency. These metrics measure whether a method produces clean and coherent videos, but they do not fully
capture whether the generated motion satisfies the prompted physical event.

Second, we use \lqr{GPT5.5} to evaluate physical plausibility and
prompt-event alignment. For each generated video, the judge receives the input
image, the event prompt, and sampled video frames, and scores the result on a
five-point scale. The judge considers whether the video preserves the intended
objects, realizes the correct contact or force direction, respects support and
gravity constraints, and avoids physically implausible object responses. We
report \textit{PhysReal} as the average GPT physical-plausibility score after
normalization to [0,1]. Additionally, we conduct a two-alternative forced-choice (2AFC) user study. Each participant evaluates six test scenes for each of the five baseline comparisons. In each trial, participants view an action description together with two side-by-side videos presented in random order: one generated by our method and one by a baseline. They select the better video according to one of three criteria: action following, visual quality, or physical plausibility.

\paragraph{Implementation details.}
All methods are run under the same evaluation protocol. When a method supports
image-conditioned generation, we provide the selected initial image together
with the original prompt. For text-only baselines, we provide the original
prompt and do not use any additional image-specific oracle information. Unless
otherwise stated, each method generates one video per case. For stochastic
methods, we use the same random seed budget and report the average score across
generated samples. Our method uses a maximum reflection budget of 10 rounds.
The loop terminates early once the diagnostic rollout passes the stage-specific
verification. All generated videos are evaluated by the same VBench and
GPT-based judging pipelines without method-specific post-processing.

\begin{figure*}[t]
    \centering
    \includegraphics[width=\textwidth]{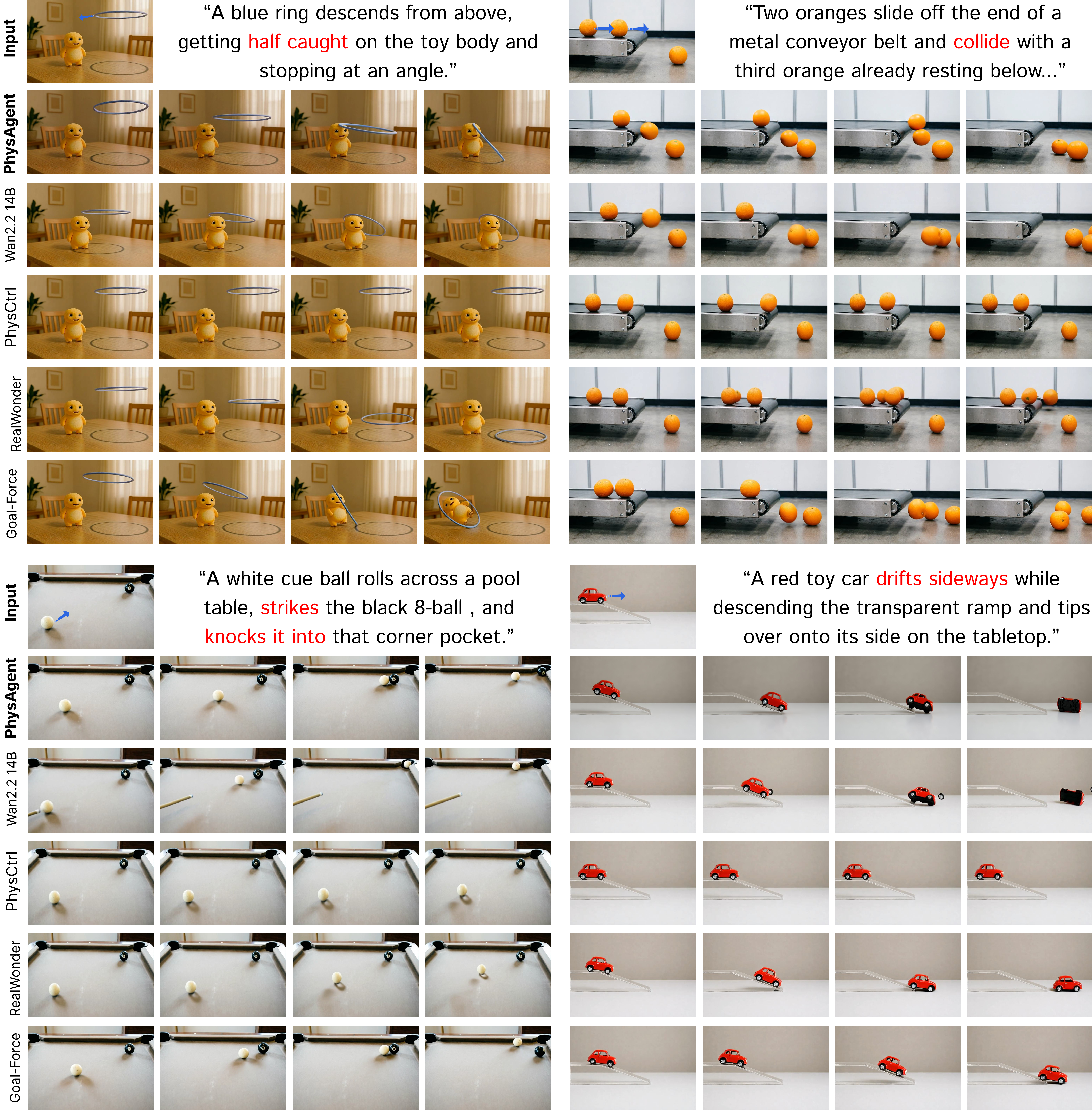}
    \caption{
        Qualitative comparison on representative physical interaction cases.
        Compared with baselines, our method better preserves the intended event
        structure, including contact timing, force direction, support
        constraints, and object response.
    }
    \vspace{-0.5in}
    \label{fig:qualitative}
\end{figure*}

\subsection{Comparison with State-of-the-Art Methods}

\begin{table}[t]
\centering
\small
\setlength{\tabcolsep}{3pt}
\caption{
Quantitative comparison to baselines. Visuals, Aesthetics, and Consistency
follow VBench-style automatic video-quality verification, while PhysReal is a
normalized GPT-based score for physical plausibility and prompt-event
alignment.
}
\label{tab:quantitative_comparison}
\begin{tabular}{llcccc}
\toprule
\textbf{Type} & \textbf{Methods} & \textbf{Visuals} ($\uparrow$) & \textbf{Aesthetics} ($\uparrow$) & \textbf{Consistency} ($\uparrow$) & \textbf{PhysReal} ($\uparrow$) \\
\midrule
\multirow{2}{*}{TI2V}
& Wan2.2 14B~\citep{wan2025wan}              & 0.625 & 0.486 & \textbf{0.274} & 0.633 \\
& Goal-Force~\citep{gillman2026goal}         & 0.634 & 0.488 & 0.262 & 0.671 \\
\midrule
\multirow{4}{*}{\shortstack{Sim.-\\based}}
& PhysCtrl~\citep{wang2025physctrl}          & 0.634 & 0.505 & 0.252 & 0.579 \\
& PhysGen3D~\citep{chen2025physgen3d}        & 0.622 & 0.474 & 0.247 & 0.480 \\
& RealWonder~\citep{liu2026realwonder}       & 0.628 & 0.493 & 0.237 & 0.631 \\
\cmidrule(l){2-6}
& \textbf{Ours}                              & \textbf{0.644} & \textbf{0.521} & \underline{0.266} & \textbf{0.714} \\
\bottomrule
\end{tabular}
\end{table}

\label{sec:sota_comparison}

\textbf{Quantitative Results:} Table~\ref{tab:quantitative_comparison} reports the quantitative comparison on our physical interaction benchmark. General text-image-to-video (TI2V) methods can often produce temporally smooth videos and may make the prompted event appear at a coarse level. However, their motion is generated by the video prior rather than by an executed physical process. As a result, they can force an event to happen visually while producing intermediate dynamics that violate physical constraints, such as incorrect contact timing, implausible force direction, missing support constraints, or material-independent object responses. This distinction explains why such methods can obtain competitive VBench-style consistency, while still receiving lower PhysReal scores. In contrast, PhysAgent explicitly constructs, executes, and verifies a physical control program before video synthesis, leading to stronger physical plausibility and prompt-event alignment.

The gap is also evident when comparing against physics-grounded simulation baselines. These methods introduce physical priors, but they often depend on a single round of attribute estimation or simulator configuration. Such one-shot estimates are not robust to errors in object grounding, material assignment, or action timing. Since these factors are strongly coupled, a locally plausible parameter choice can still lead to a rollout that fails the intended event. PhysAgent addresses this issue by treating each physical program as an executable hypothesis: it runs a diagnostic simulation, checks the resulting motion, and revises only the failed factors through targeted feedback. This closed-loop process is especially beneficial for events that require delayed release, directed impact, state-dependent motion, or object-to-object targeting.

\textbf{\lqr{User study.}}  We report quantitative user study results in Table~\ref{tab:human_2afc}
using the 2AFC setup described above. PhysAgent is consistently preferred over all baseline methods.


\begin{table}[t]
\centering
\caption{
2AFC human study results. Each entry reports the favor rate of PhysAgent over the corresponding baseline method.
}
\label{tab:human_2afc}
\resizebox{\linewidth}{!}{%
\begin{tabular}{lccc}
\toprule
\textbf{Comparison} &
\textbf{Action Following} &
\textbf{Visual Quality} &
\textbf{Physical Plausibility} \\
\midrule
over Wan2.2 14B~\cite{wan2025wan}        & 66.8\% & 60.7\% & 73.4\% \\
over Goal-Force~\cite{gillman2026goal}   & 70.2\% & 64.8\% & 66.5\% \\
over RealWonder~\cite{liu2026realwonder} & 74.9\% & 69.6\% & 75.2\% \\
over PhysCtrl~\cite{wang2025physctrl}    & 90.8\% & 88.9\% & 92.1\% \\
over PhysGen3D~\cite{chen2025physgen3d}  & 92.4\% & 90.3\% & 93.6\% \\
\bottomrule
\end{tabular}%
}
\vspace{-0.6em}
\end{table}

\subsection{Ablation Studies}

\begin{table}[t]
    \centering
    \setlength{\tabcolsep}{4.5pt}
    \caption{
        Ablation study on the physical interaction benchmark. We evaluate the
        contribution of the reflection loop and semantic control APIs using the
        same automatic video-quality metrics.
    }
    \label{tab:ablation}
    \begin{tabular}{lccccc}
        \toprule
        Variant & Visuals $\uparrow$ & Aesthetics $\uparrow$ & Consistency $\uparrow$ & PhysReal $\uparrow$ & Avg. Rounds \\
        \midrule
        w/o reflection loop 
            & 0.628 & 0.493 & 0.237 & 0.631 & 1.00 \\
        w/o semantic control APIs 
            & 0.636 & 0.514 & 0.251 & 0.667 & 5.74 \\
        \midrule
        \textbf{Ours (Full)}
            & \textbf{0.644} & \textbf{0.521} & \textbf{0.266} & \textbf{0.714} & 3.63 \\
        \bottomrule
    \end{tabular}
\end{table}

\begin{figure}[t]
    \centering
    \includegraphics[width=\textwidth]{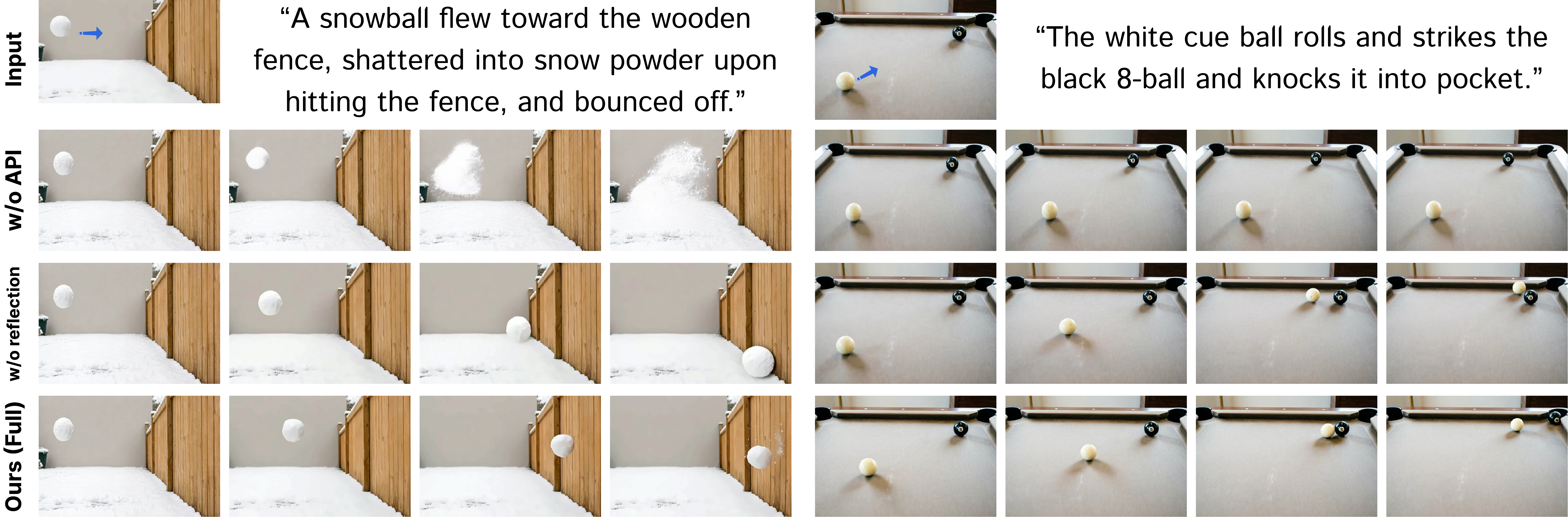}
    \caption{
        Ablation for the core components of our method.
    }
    \label{fig: ablation}
\end{figure}

We conduct ablation studies to isolate the contribution of each component in
our framework. Removing the reflection loop reduces the system to a one-shot
physics configuration pipeline, which makes it difficult to correct wrong object
selection, implausible support placement, or misdirected actuation. Replacing the semantic control APIs with low-level
force parameters reduces the expressiveness of the event program and makes
complex interactions more brittle.

As shown in Table~\ref{tab:ablation} and Figure~\ref{fig: ablation}, both components contribute to the final performance. Removing the reflection loop causes the largest drop in PhysReal, confirming that one-shot simulator configurations are difficult to correct once they fail. Replacing the semantic control APIs with low-level parameters also reduces all evaluation scores and increases the average number of refinement rounds from 3.63 to 5.74. This result suggests that the semantic APIs improve not only control expressiveness but also refinement efficiency.

\subsection{Different Actions on the Same Scene}

\begin{figure}[t]
    \centering
    \includegraphics[width=1.0\textwidth]{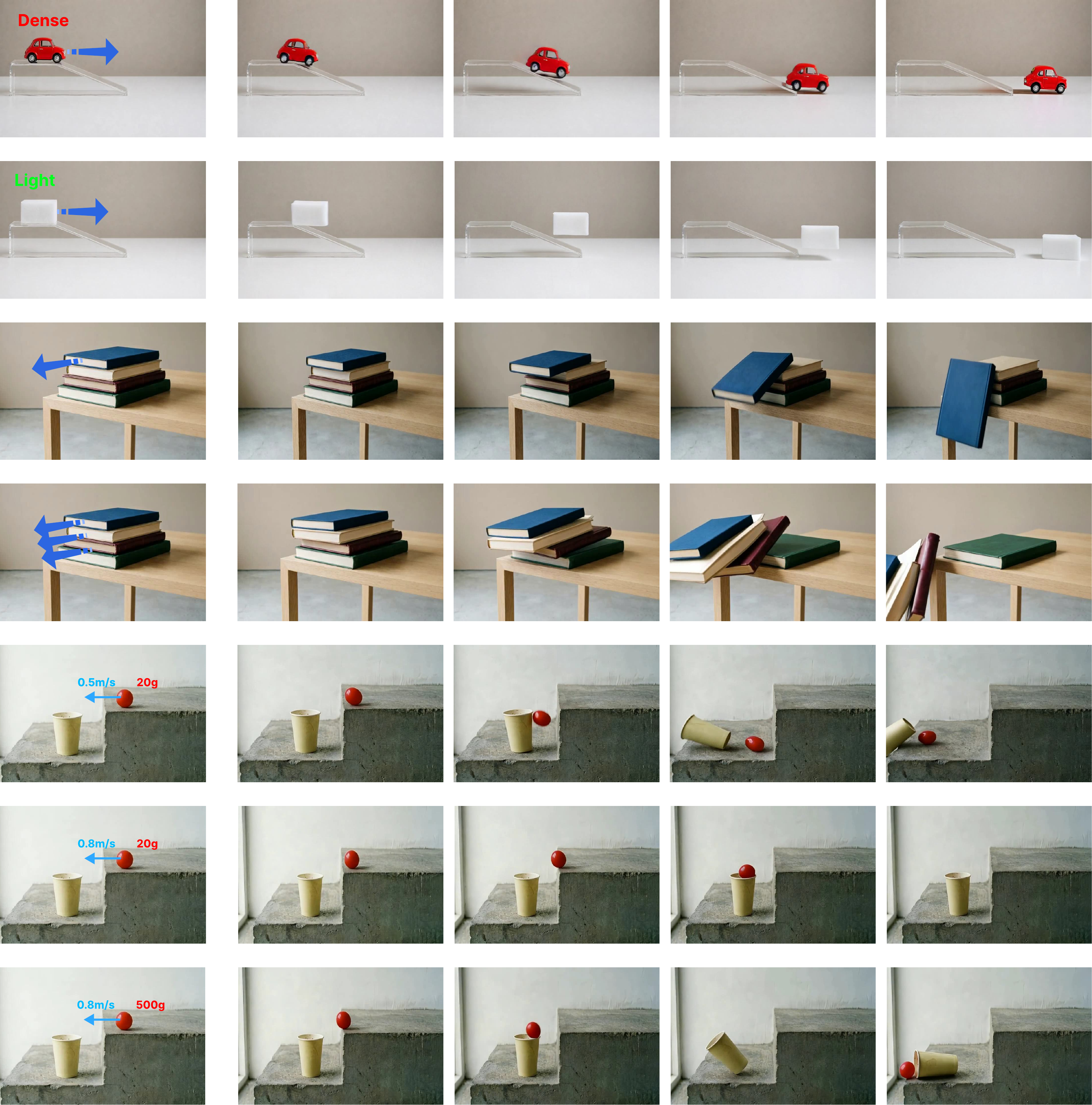}
    \caption{
        Different events on the same scene. By changing the physical material or interaction parameters while keeping the visual scene fixed, our method produces different motion responses that reflect the intended
        physical properties.
    }
    \label{fig: different event}
\end{figure}

Having established the contribution of the reflection loop and semantic control APIs, we next examine whether the framework can realize distinct physical outcomes while preserving the same visual scene. Given the same input image, the
agent can instantiate different event programs by changing only the semantic
control primitives, such as the force direction, force duration, release timing,
torque magnitude, fixed object, or target object. This allows the generated videos to preserve the same scene identity while realizing distinct physical outcomes. Examples include pushing an object to the left or right, releasing it before or after contact, rotating it instead of translating it, and directing it toward different targets. Results are shown in Figure~\ref{fig: different event}.

\section{\lqr{Conclusion}}
\label{sec:conclusion}

We presented PhysAgent, a reflective agentic physics-control framework for physically plausible video generation. Rather than treating physical configuration as a one-shot prediction, PhysAgent formulates it as an executable control program that can be simulated, evaluated, and revised before video synthesis. The framework separates scene state from event dynamics and progressively validates intermediate stages. Its semantic physics-control APIs provide a structured interface for expressing event-level motion, thereby reducing the brittleness of low-level force specifications. Experiments show that PhysAgent improves physical plausibility and prompt-event alignment over both general TI2V methods and physics-grounded baselines. The results further demonstrate that iterative feedback and semantic control APIs are complementary: feedback enables the agent to correct reconstruction and simulation failures, whereas structured APIs make physical actuation easier to express and revise. These findings suggest that execution-aware physical control is a promising direction for controllable video generation, particularly when prompts require precise contact timing, force direction, material response, and multi-stage object interactions.

\begin{figure}[t]
    \centering
    \includegraphics[width=1.1\textwidth]{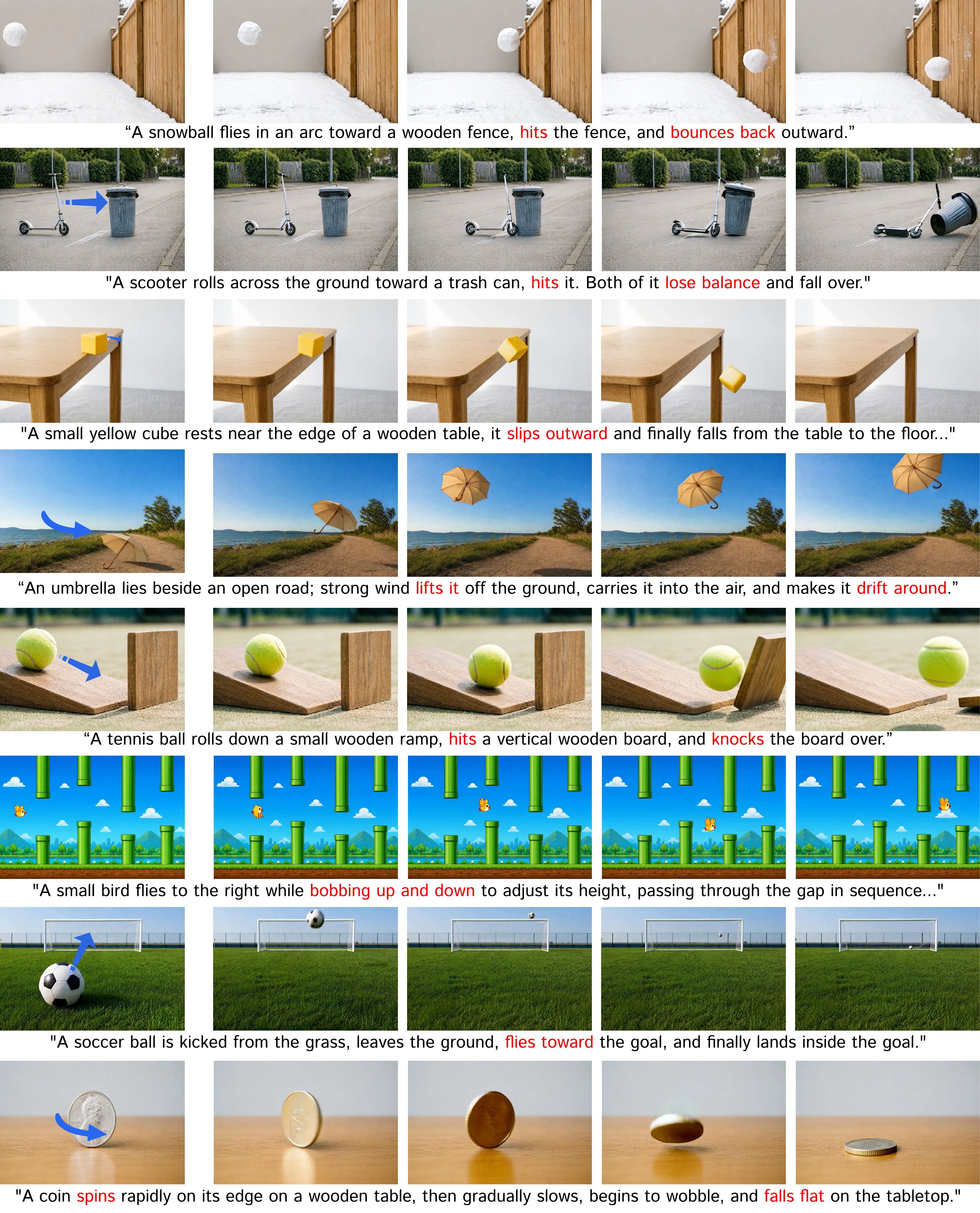}
    \caption{
        Qualitative Result of PhysAgent.
    }
    \label{fig:qualitative_fig}
\end{figure}

\clearpage

\bibliographystyle{plainnat}
\bibliography{references}

\end{document}